%% file: main.tex
\documentclass[10pt,reqno,cmintegrals,cmbraces]{amsproc}
\usepackage{titlesec}
\UseRawInputEncoding
\usepackage{newtxtext}
\usepackage{bbold}
\usepackage{anyfontsize}
\usepackage[cmintegrals,cmbraces]{newtxmath}
\titleformat{\section}
{\normalfont\Large\bfseries}{\thesection}{1em}{}
\titleformat{\subsection}
{\normalfont\large\bfseries}{\thesubsection}{1em}{}
\titleformat{\subsubsection}
{\normalfont\normalsize\bfseries}{\thesubsubsection}{1em}{}
\titleformat{\paragraph}[runin]
{\normalfont\normalsize\bfseries}{\theparagraph}{1em}{}
\titleformat{\subparagraph}[runin]
{\normalfont\normalsize\bfseries}{\thesubparagraph}{1em}{}
\usepackage{booktabs} 
\usepackage[colorinlistoftodos]{todonotes}
\usepackage{cite}
\usepackage{color}
\definecolor{RED}{HTML}{FF0000}
\usepackage{graphicx}
\usepackage{psfrag,epsfig}
\usepackage{mathrsfs}
\usepackage{tabularx}
\usepackage{xr}
\usepackage{arydshln}
\usepackage{siunitx,array,multirow}
\usepackage{rotating}
\usepackage[letterpaper,margin=1in]{geometry}
\usepackage{subfigure}
\usepackage{enumerate}
\usepackage{comment}
\usepackage{lineno}
\usepackage{algorithmic}
\usepackage{algorithm}
\usepackage[small]{caption}
\usepackage{bbold}
\usepackage[normalem]{ulem}
\usepackage{booktabs}
\usepackage{zref-user}
\usepackage[debug=false, colorlinks=true, pdfstartview=FitV, linkcolor=blue, citecolor=blue, urlcolor=blue]{hyperref}
\hypersetup{
    colorlinks=true,
    linkcolor=blue,
    filecolor=magenta,      
    urlcolor=blue,
}
\usepackage{float}
\mathchardef\mhyphen="2D
\usepackage{chemarr}
\usepackage[version=3]{mhchem}
\newcommand*\patchAmsMathEnvironmentForLineno[1]{%
  \expandafter\let\csname old#1\expandafter\endcsname\csname #1\endcsname
  \expandafter\let\csname oldend#1\expandafter\endcsname\csname end#1\endcsname
  \renewenvironment{#1}%
     {\linenomath\csname old#1\endcsname}%
     {\csname oldend#1\endcsname\endlinenomath}}%
\newcommand*\patchBothAmsMathEnvironmentsForLineno[1]{%
  \patchAmsMathEnvironmentForLineno{#1}%
  \patchAmsMathEnvironmentForLineno{#1*}}%
\AtBeginDocument{%
\patchBothAmsMathEnvironmentsForLineno{equation}%
\patchBothAmsMathEnvironmentsForLineno{align}%
\patchBothAmsMathEnvironmentsForLineno{flalign}%
\patchBothAmsMathEnvironmentsForLineno{alignat}%
\patchBothAmsMathEnvironmentsForLineno{gather}%
\patchBothAmsMathEnvironmentsForLineno{multline}%
}

\title{HyDeMiC: A Deep Learning-based Mineral Classifier using Hyperspectral Data}

\author[Mamud et al.,]
{M.~L.~Mamud$^{*,1}$, 
Piyoosh~Jaysaval$^{*,1}$,
Frederick~D~Day-Lewis$^1$, and
M.~K.~Mudunuru$^1$\\
{\small $^1$Subsurface Science Group, Pacific Northwest National Laboratory, Richland, WA 99352.\\ 
$^*$Corresponding authors:Md Lal Mamud (~\texttt{lal.mamud@pnnl.gov}) and\\ Piyoosh~Jaysaval (~\texttt{Piyoosh.Jaysaval@pnnl.gov}).\\
PNNL-SA-215088
}}

\date{\today}

\begin{document}
\maketitle

\raggedbottom

\input{sections/abstract}

\input{sections/s1_intro}
\input{sections/s2_methods}
\input{sections/s3_results}
\input{sections/s4_conclusion}

\section*{Acknowledgment and Disclaimer}
The U.S. Department of Energy supported this work under grant no. FWP 81034.
This research work was prepared as an account of work sponsored by an agency of the United States Government. 
Neither the United States Government nor any agency thereof, nor any of their employees, makes any warranty, express or implied, or assumes any legal liability or responsibility for the accuracy, completeness, or usefulness of any information, apparatus, product, or process disclosed, or represents that its use would not infringe privately owned rights. 
Reference herein to any specific commercial product, process, or service by trade name, trademark, manufacturer, or otherwise does not necessarily constitute or imply its endorsement, recommendation, or favoring by the United States Government or any agency thereof. 
The views and opinions of authors expressed herein do not necessarily state or reflect those of the United States Government or any agency thereof.

\bibliographystyle{IEEEtran}
\bibliography{references}

\end{document}

%% file: sections/abstract.tex
\textbf{Abstract:}~
Hyperspectral imaging (HSI) has emerged as a powerful remote sensing tool for mineral exploration, capitalizing on unique spectral signatures of minerals. 
However, traditional classification methods such as discriminant analysis, logistic regression, and support vector machines often struggle with environmental noise in data, sensor limitations, and the computational complexity of analyzing high-dimensional HSI data. This study presents HyDeMiC (Hyperspectral Deep Learning-based Mineral Classifier), a convolutional neural network (CNN) model designed for robust mineral classification under noisy data. To train HyDeMiC, laboratory-measured hyperspectral data for 115 minerals spanning various mineral groups were used from the United States Geological Survey (USGS) library. The training dataset was generated by convolving reference mineral spectra with an HSI sensor response function. 
These datasets contained three copper-bearing minerals (i.e, Cuprite, Malachite, and Chal-copyrite) used as case studies for performance demonstration. The trained CNN model was evaluated on several synthetic 2D hyperspectral datasets with a range of noise levels to replicate realistic field conditions. 
The HyDeMiC's performance was assessed using the Matthews Correlation Coefficient (MCC), providing a comprehensive measure across different noise regimes. Results demonstrate that HyDeMiC achieved near-perfect classification accuracy (MCC = 1.00) on clean and low-noise datasets and maintained strong performance under moderate noise conditions. 
These findings emphasize HyDeMiC's robustness in the presence of moderate noise, highlighting its potential for real-world applications in hyperspectral imaging, where noise is often a significant challenge.

\noindent\textbf{Keywords:}~
Hyperspectral imaging (HSI), 
Mineral classification, 
Critical material, 
Convolutional Neural Network (CNN), 
Deep learning, 
USGS Spectral Library.

%% file: sections/s1_intro.tex
\section{Introduction}

Hyperspectral imaging (HSI) has become an essential tool in remote sensing applications, providing detailed spectral information across a wide range of wavelengths. 
This rich spectral information enables the identification and classification of materials based on their unique spectral signatures \cite{Goetz2009ThreeView}. 
In geoscience, HSI is particularly valuable for mineral exploration, as different minerals exhibit distinct spectral reflectance properties across the electromagnetic spectrum \cite{Peyghambari2021HyperspectralReview}. 
Accurate mineral classification using hyperspectral data is crucial for understanding the Earth's surface composition and has widespread applications in mining, subsurface monitoring, and resource management \cite{Fasnacht2019ANm}.

Traditional mineral classification methods are based on manual interpretation of hyperspectral data coupled with conventional machine learning techniques such as support vector machines and decision trees \cite{Asadzadeh2016ASensing}. 
However, these methods struggle with the high dimensionality and complexity of hyperspectral datasets, leading to increased computational costs and reduced classification accuracy \cite{Paoletti2019DeepReview}. 
Additionally, modern hyperspectral sensors generate vast amounts of data, making manual analysis prone to human error, impractical, or intractable. 
To address these challenges, deep learning techniques such as Convolutional Neural Networks (CNNs) have emerged as a powerful alternative for automating mineral classification from hyperspectral data \cite{BenHamida20183-DClassification, Li2019DeepOverview, Chen2022HyperspectralLibrary}.

CNNs have demonstrated remarkable success in hyperspectral image classification by effectively capturing spatial and spectral features \cite{Yu2017ConvolutionalClassification, Zhang2020ClassificationNetwork}. 
Their hierarchical feature-learning capability enables them to outperform traditional classification methods in accuracy and robustness. 
However, noise contamination remains a significant challenge in hyperspectral mineral classification. 
Noise in hyperspectral data can arise from multiple sources, including sensor data collection inaccuracies, environmental conditions, and atmospheric interference, potentially distorting spectral signatures and leading to misclassification \cite{Hang2020ClassificationCNNs, Fricker2019AImagery}.

Since real-world hyperspectral data is rarely noise-free, developing deep learning models that are robust to noise is critical while maintaining high classification accuracy. 
To this end, we present HyDeMiC (\textit{Hyperspectral Deep Learning-based Mineral Classifier}), a trained CNN-based framework designed for mineral classification under noisy conditions. 
The model comprises of multiple convolutional and pooling layers, with dropout and regularization techniques employed to mitigate overfitting. 
HyDeMiC was trained on simulated hyperspectral signatures of 115 minerals from the United States Geological Survey (USGS) spectral library \cite{Kokaly2017USGSRelease} using a supervised learning approach, optimizing classification accuracy by minimizing a categorical cross-entropy loss function. 



To evaluate HyDeMiC’s performance under controlled yet challenging conditions, we tested the trained model using a synthetic 2D hyperspectral dataset containing three copper-bearing minerals: Cuprite, Malachite, and Chalcopyrite. These minerals were selected due to their distinct spectral signatures and relevance to real-world exploration scenarios because the U.S. Department of Energy designated them as critical materials \cite {U.S.DepartmentofEnergy2023CriticalAssessment}. 
By systematically analyzing classification accuracy across varying noise levels, we demonstrate HyDeMiC’s effectiveness in handling noisy hyperspectral datasets, highlighting its potential for real-world mineral exploration applications.

%% file: sections/s2_methods.tex
\section{Methodology}

\subsection{Training data for HyDeMiC}
To construct a diverse and representative training dataset, we used 444 unique 1D hyperspectral reflectance spectra from the USGS Mineral Spectral Library \cite{Kokaly2017USGSRelease}, representing 115 distinct minerals. 
These minerals belong to a wide range of mineral groups, including silicates, oxides, carbonates, sulfates, sulfides, halides, and borates, as well as hydroxides, phosphates, and mineraloids, ensuring comprehensive spectral variability in the dataset. 
Each spectrum was processed to simulate the hyperspectral response of the Airborne Visible/Infrared Imaging Spectrometer (AVIRIS) system, resulting in 224 spectral bands spanning a broad range of wavelengths.
AVIRIS is a well-established hyperspectral sensor known for its high spectral resolution and extensive application in geological and mineralogical remote sensing. 
These datasets originate from laboratory-based measurements of mineral samples, ensuring high spectral fidelity and accurate spectral representation. 
The spectral data we used had already been preprocessed in the USGS library. 
Specifically, the raw mineral spectra were convolved with the AVIRIS sensor response function. 
This process incorporates AVIRIS's spectral sensitivity and bandpass characteristics, transforming high-resolution laboratory spectra into a dataset that closely mimics real-world airborne hyperspectral observations. 

\subsection{Architecture of HyDeMiC}
\begin{figure*}[!t]
\centering
    \includegraphics[width=\textwidth]{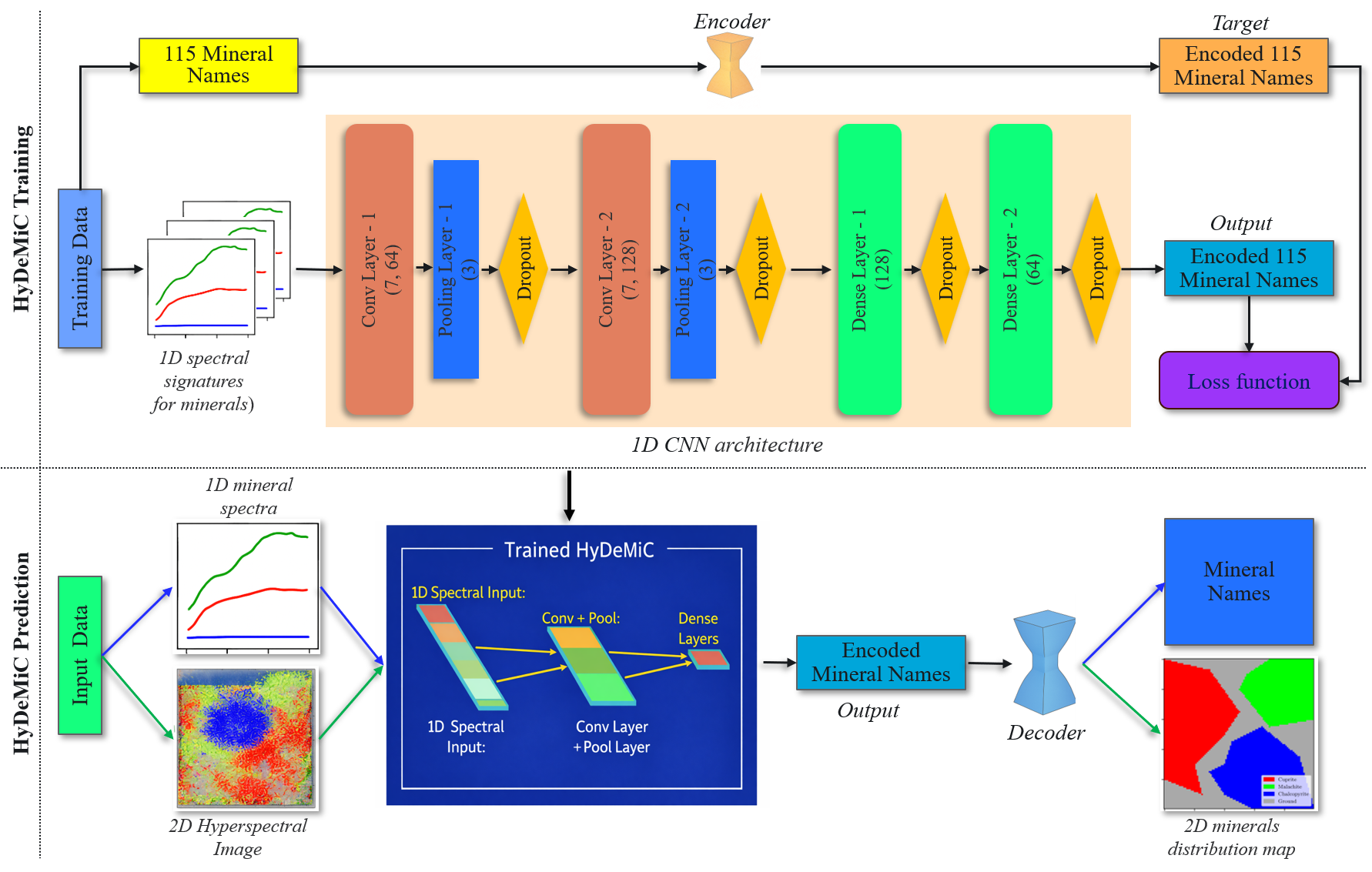}
    \caption{Figure: HyDeMiC framework for mineral classification using hyperspectral data and a 1D Convolutional Neural Network (CNN) architecture.
The top section illustrates the training workflow, in which 1D spectral signatures from minerals are processed by a CNN comprising two convolutional layers (with kernel size 7 and increasing filter depth), max-pooling, dropout for regularization, and fully connected layers. The model is trained to predict encoded labels of 115 mineral types using an encoder-decoder strategy and a loss function computed against the encoded target labels.
The bottom section depicts the prediction workflow, in which the trained 1D CNN model takes 1D spectral signatures or 2D hyperspectral images to generate encoded mineral predictions. These encoded outputs are decoded to produce human-readable mineral names, which are subsequently visualized as a 2D mineral distribution map.}
\label{fig:cnn_archietecture}
\end{figure*}

The HyDeMiC model architecture used in this study was designed to classify hyperspectral data under varying noise levels. 
It consists of multiple convolutional, pooling, and fully connected layers, enabling efficient extraction of spectral features from 1D hyperspectral data (Figure \ref{fig:cnn_archietecture}). 
The model ingest simulated 1D spectral signatures as input and provides the corresponding mineral classification as output (Figure \ref{fig:cnn_archietecture}). 
The architecture begins with two sequential 1D convolutional layers, each with a kernel size of 7. 
The number of filters increases from 64 in the first layer to 128 in the second, enabling the network to progressively extract more complex spectral features. 
Each convolutional layer is followed by batch normalization to stabilize learning, a LeakyReLU activation to introduce nonlinearity, and a max-pooling layer with a pool size of 3 to reduce dimensionality while retaining important features. (Figure~\ref{fig:cnn_archietecture}). 
Dropout layers are applied after both pooling operations to improve generalization and reduce overfitting. 
The resulting feature maps are flattened and passed through two fully connected (dense) layers with 128 and 64 units, respectively, followed by LeakyReLU activations and dropout. 
Finally, a dense output layer with 116 units corresponding to the number of mineral classes is used for classification. 
This layered structure, combined with the use of dropout, batch normalization, and non-linear activations, enables the model to effectively extract, refine, and interpret spectral information across a wide range of mineral types and noise scenarios (Figure \ref{fig:cnn_archietecture}).

The trained HyDeMiC model operates in prediction mode, as shown in the lower panel of Figure~\ref{fig:cnn_archietecture}.
At this stage, the trained 1D CNN processes unseen hyperspectral inputs, which may be individual 1D spectral signatures or pixel-wise spectra from 2D hyperspectral images.
Each input spectrum is processed independently through the learned convolutional and fully connected layers to generate an encoded mineral representation.

The encoded outputs are then passed through a decoder that maps latent representations to human-interpretable mineral-class labels. For 2D hyperspectral images, this process is applied to each pixel spectrum, enabling reconstruction of a spatially resolved mineral classification map. The resulting 2D mineral distribution map preserves the spatial organization of the original image and assigns a mineral label to each pixel, bridging spectral-level learning and image-level interpretation.

\subsection{Training of HyDeMiC}
The HyDeMiC model was trained on a one-dimensional hyperspectral dataset of 444 distinct spectral signatures representing 115 minerals. 
Each signature comprises 224 bands, capturing the reflectance characteristics of individual minerals.
These minerals belong to various mineral groups, including silicates, oxides, carbonates, sulfates, sulfides, halides, borates, hydroxides, phosphates, and mineraloids, ensuring a diverse and representative spectral dataset. 
The HyDeMiC model aims to classify the hyperspectral data into 115 mineral categories with high accuracy and robustness across different mineral groups. 
The model was hyperparameter-tuned through a series of iterative experiments to achieve optimal performance. 
The training was carried out for 2,000 epochs with a batch size of 16, using an initial learning rate of $1\times 10^{-3}$, which was dynamically adjusted with a cosine annealing learning rate scheduler. 
This method allowed for a gradual decline in the learning rate, promoting stable convergence and preventing premature saturation. 
Optimization was improved by employing the AdamW optimizer, which incorporates weight decay to enhance generalization, particularly for high-dimensional spectral data \cite{Loshchilov2017DecoupledRegularization}. 
Since the classification problem involved multiple mineral classes, categorical cross-entropy was used as the loss function to minimize the divergence between the predicted and true class distributions.

The categorical cross-entropy loss function for a multi-class classification problem is defined as \cite{Goodfellow2016DeepLearning}:
\begin{equation}
\text{Loss} = -\sum_{i=1}^{N} \sum_{c=1}^{C} y_{i,c} \log(\hat{y}_{i,c})
\label{eq:cross_entropy}
\end{equation}
where \( N \) is the number of samples, \( C \) is the number of classes, \( y_{i,c} \) is a binary indicator (0 or 1) denoting whether class label \( c \) is the correct classification for sample \( i \), and \( \hat{y}_{i,c} \) is the predicted probability of sample \( i \) being in class \( c \).

By leveraging a well-curated, sensor-calibrated dataset, HyDeMiC was trained to classify minerals based on their hyperspectral signatures robustly. The model’s training strategy ensures its applicability to real-world geological and remote sensing scenarios, where accurate mineral identification is critical for exploration, resource management, and environmental monitoring.

\subsection{Performance evaluation metrics}
To evaluate the trained HyDeMiC's predictive performance, we used the Matthews Correlation Coefficient (MCC), True Positive Rate (TPR), and Prediction Confidence (PC).
The MCC for multiclass classification is a straightforward extension to the MCC commonly used for binary classification. It measures classification quality, considering true and false positives and negatives across multiple classes. 
The MCC in the multiclass case is given by \cite{Matthews1975ComparisonLysozyme}:
\begin{equation}
    \text{MCC} = \frac{\sum_{k} \sum_{l} \sum_{m} M_{kk} M_{lm} - M_{kl} M_{mk}}{\sqrt{\sum_{k} \left( \sum_{l} M_{kl} \right) \left( \sum_{l} M_{lk} \right) \sum_{k} \left( \sum_{l} M_{lk} \right) \left( \sum_{l} M_{kl} \right)}}
    \label{eq:mcc}
\end{equation}
where \( M_{kk} \) is the number of instances correctly predicted as class \( k \), \( M_{lm} \) represents instances belonging to true class \( l \) but misclassified as class \( m \), \( M_{kl} \) is the number of instances with true label \( k \) but predicted as class \( l \), and \( M_{mk} \) represents instances with true label \( m \) but predicted as \( k \).

The True Positive Rate (TPR), also called Sensitivity or Recall, quantifies the model’s ability to identify positive samples correctly. It is defined as follows\cite{Powers2011Evaluation:Correlation}: 
\begin{equation}
\text{TPR} = \frac{\sum\limits_{i=1}^{N} \mathbf{1}(\hat{y}_i = 1\ \textit{and}\ y_i = 1)}{\sum\limits_{i=1}^{N} \mathbf{1}(y_i = 1)}
\label{eq:tpr}
\end{equation}
where \( N \) is the number of samples, and \( \mathbf{1} \) is the indicator function, which is 1 if both the predicted class \( \hat{y}_i \) and the true class \( y_i \) are positive, and 0 otherwise. The denominator represents the total number of actual positive samples in the dataset.

The prediction confidence (PC) \( c_i \) for sample \( i \) is calculated as \cite{Goodfellow2016DeepLearning}:
\begin{equation}
c_i = \max(\hat{y}_i^{(1)}) \times 100
\label{eq:confidence}
\end{equation}
where \( \hat{y}_i^{(1)} \) represents the predicted probability for the most likely class. Confidence is expressed as a percentage, with higher values indicating more substantial confidence in the prediction.

To further analyze the reliability of the predicted confidence scores, we examined the empirical distribution of prediction confidence for correct and incorrect classifications using density-normalized histograms.
For a given confidence bin $i$, the probability density $p(c_i)$ was computed as
\begin{equation}
p(c_i) = \frac{N_i}{N \, \Delta c},
\label{eq:confidence_density}
\end{equation}
where $N_i$ denotes the number of samples whose prediction confidence falls within bin $i$, 
$N$ is the total number of samples in the corresponding group (correct or incorrect predictions), 
and $\Delta c$ is the bin width \cite{Scott1992MultivariateVisualization}.
This normalization ensures that the area under each confidence distribution integrates to unity,
\begin{equation}
\int p(c)\, dc = 1,
\end{equation}
allowing direct comparison of confidence distributions between correct and incorrect predictions, even in the presence of class imbalance.
Histogram bin widths were determined using the Freedman--Diaconis rule \cite{Freedman1981OnTheory} and capped to ensure visual clarity across different noise levels.

\subsection{Mineral classification with HyDeMiC}
\begin{figure*}[!t]
\centering
    \includegraphics[width=\textwidth]{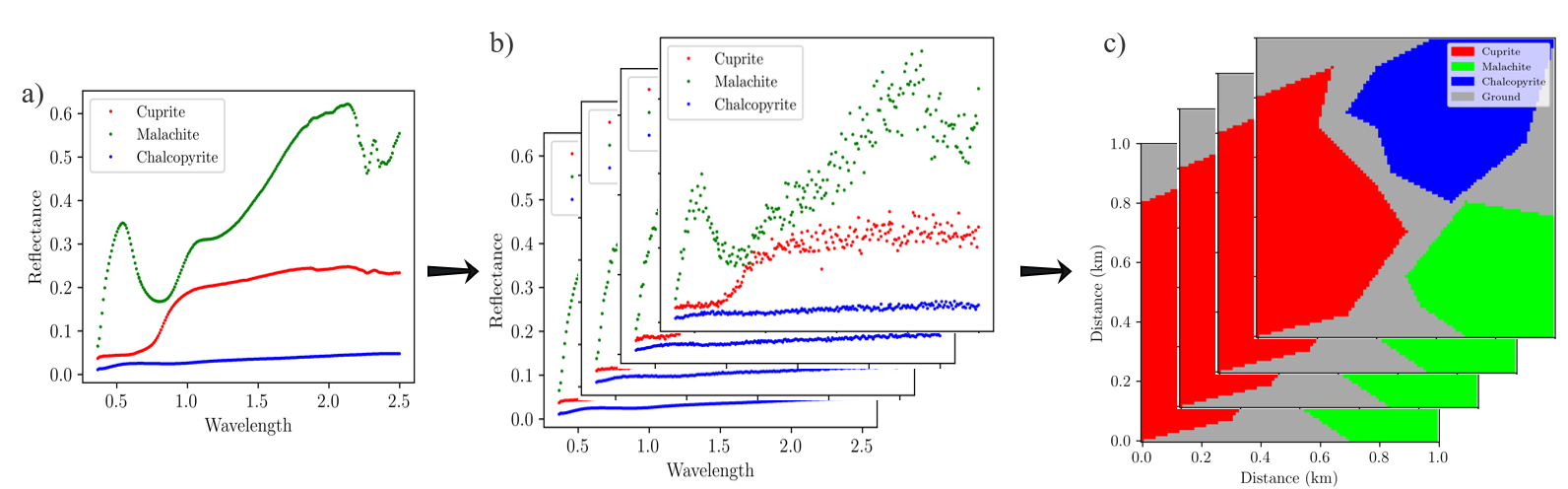}
    \caption{a) 1D reflectance spectral signature of Cuprite, Malachite, and Chalcopyrite, b) Generation of 1D synthetic spectral data with 1-10\% of random noise added to the 1D spectral signature for the selected minerals, and c) schematic illustration of 2D synthetic hyperspectral images created using 1D synthetic spectral data by assigning them to each pixel for evaluating HyDeMiC's prediction capability.}
\label{fig:synthetic_data_generation}
\end{figure*}

we created 2D synthetic hyperspectral images with a spatial resolution of $100 \times 100$ pixels, representing a 1 km$^2$ area to apply and evaluate the performance of the HyDeMiC. 
To create 2D synthetic hyperspectral images, we selected 1D reflectance spectral signatures for three copper-bearing minerals of interest, as discussed in the introduction (Figure \ref{fig:synthetic_data_generation}a). 
We created a segmented 2D spatial domain structured into distinct polygonal regions corresponding to these three minerals and the non-mineralized (ground) region. 
This structured approach ensured a geologically realistic spatial distribution of mineralized and non-mineralized regions.

To simulate real-world conditions, we generated 1D synthetic reflectance spectral data by introducing controlled levels of random Gaussian noise (1\%, 2\%, 5\%, and 10\%) into the baseline 1D spectral signatures of Cuprite, Malachite, and Chalcopyrite (Figure \ref{fig:synthetic_data_generation}b). 
These baseline spectral signatures, shown in Figure \ref{fig:synthetic_data_generation}a, represent characteristic mineral reflectance profiles. 
%
For each noise level, 1D synthetic spectral data were assigned to individual pixels within predefined polygonal regions representing Cuprite, Malachite, Chalcopyrite, and a non-mineralized background (Figure \ref{fig:synthetic_data_generation}c). 
The non-mineralized region was assigned a uniform low-reflectance value to approximate the spectral characteristics of the natural background terrain. 
This process generated a 2D synthetic hyperspectral image that preserves spatial coherence while incorporating spectral variability. 

By progressively increasing noise levels and generating multiple 2D synthetic hyperspectral images, we created increasingly challenging conditions to assess the robustness of the HyDeMiC model. 
This approach replicates common environmental disturbances and sensor-induced variations encountered in real-world hyperspectral imaging applications. 
Evaluating the model’s performance under such conditions provides insights into its reliability and applicability for mineral classification in complex and noisy spectral environments.

%% file: sections/s3_results.tex
\section{Results and Discussion}
\subsection{HyDeMiC model training}
\begin{figure}[!t]
\centering
    \includegraphics[width=3.2in]{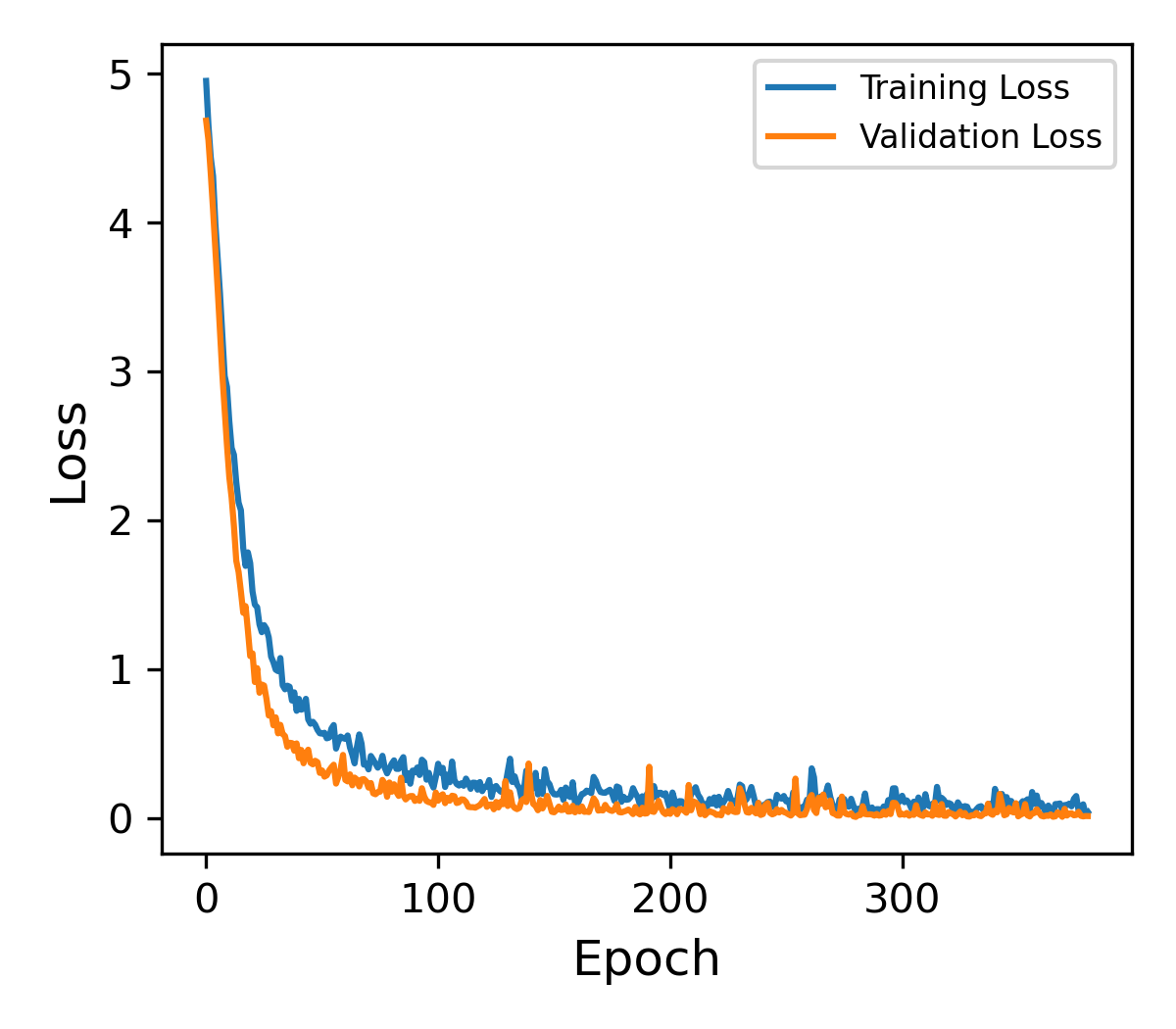}
    \caption{Training and validation loss for the developed HyDeMiC model.}
\label{fig:training_loss}
\end{figure}

HyDeMiC demonstrates stable and efficient convergence when trained on 444 unique 1D spectral reflectance signatures (Figure~\ref{fig:training_loss}). Training and validation losses decrease rapidly in the initial epochs, indicating effective early learning of spectral features. This is followed by a gradual reduction in loss, reflecting ongoing refinement of the model’s representations.

The training loss reaches a minimum of $1.29 \times 10^{-2}$ at epoch 366, while the validation loss reaches $6.94 \times 10^{-3}$ at epoch 281. The close alignment of these curves throughout training indicates strong generalization and minimal overfitting. The consistently lower validation loss is likely due to regularization and the stochastic effects of batch-wise optimization.

Minor oscillations in both loss curves appear in later epochs, as expected due to spectral similarity among mineral classes and variability in hyperspectral data. These fluctuations do not cause divergence or instability, indicating robust optimization.

Overall, the smooth convergence and consistent relationship between training and validation losses show that HyDeMiC learns compact, transferable spectral representations. This stable training supports strong predictive performance on both clean and noisy hyperspectral datasets, as discussed in the following sections.

\subsection{HyDeMiC prediction on clean hyperspectral images}
HyDeMiC delivers strong predictive performance on clean hyperspectral data, accurately classifying all 115 mineral classes used in training. This outcome demonstrates the model’s capacity to learn robust spectral representations and generalize across diverse mineralogical signatures.

\begin{figure*}[!t]
\centering
    \includegraphics[width=\textwidth]{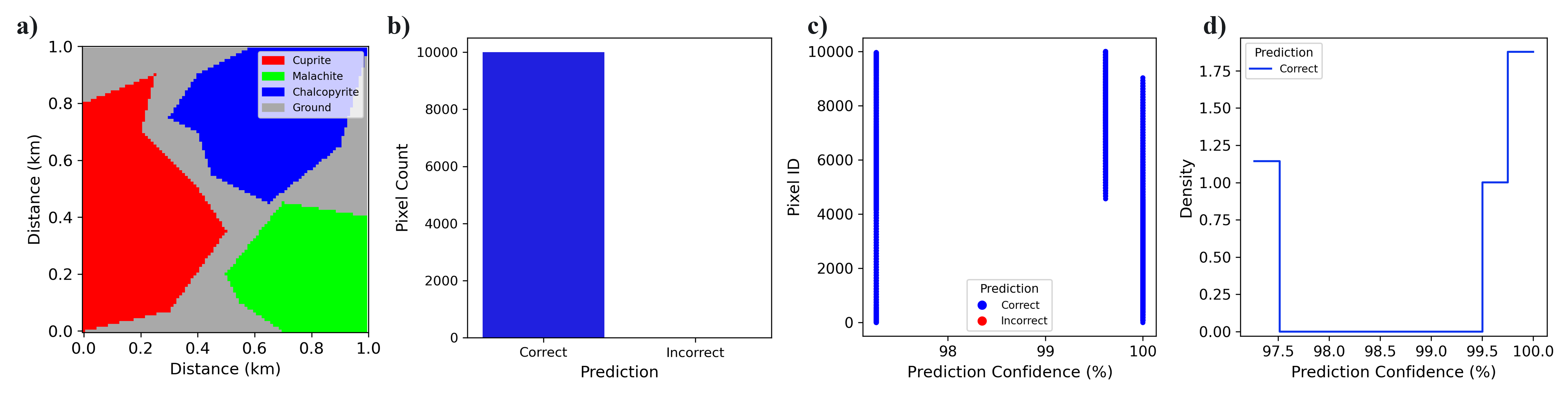}
\caption{
    HyDeMiC prediction results on clean synthetic hyperspectral data.
    (a) Spatial distribution of mineral classes predicted by HyDeMiC, 
    (b) Pixel-wise classification summary showing the total number of correct and incorrect predictions,
    (c) Prediction confidence as a function of pixel index, where each point represents an individual pixel colored by correctness, and
    (d) Density-normalized histogram of prediction confidence.
    }
\label{fig:prediction_on_clean_data}
\end{figure*}

We analyze a representative 2D synthetic hyperspectral test image containing three copper-bearing minerals: Cuprite, Malachite, and Chalcopyrite, embedded in a ground background. The mineral distribution predicted by HyDeMiC, shown in Figure~\ref{fig:prediction_on_clean_data}a, closely matches the true layout in Figure~\ref{fig:synthetic_data_generation}c, with clear mineral boundaries and no visible misclassification.

Quantitative metrics support this finding. On the noise-free dataset, HyDeMiC achieves a Matthews Correlation Coefficient (MCC) of 1.00 and a True Positive Rate (TPR) of 1.00 (Table~\ref{table_performance}), indicating perfect agreement between predicted and true labels. The pixel-wise classification summary in Figure~\ref{fig:prediction_on_clean_data}b confirms that all pixels are correctly classified, with no errors.

Prediction confidence analysis offers further insight into model reliability. As shown in Figure~\ref{fig:prediction_on_clean_data}c, confidence values for individual pixels remain consistently high across the spatial domain. The density-normalized confidence distribution in Figure~\ref{fig:prediction_on_clean_data}d shows most confidence scores are near 100\%, with a mean of 99.12\% and a median of 99.61\% (Table~\ref{table_performance}). This narrow spread indicates stable and well-calibrated predictions under ideal, noise-free conditions.

Overall, these results show that HyDeMiC effectively distinguishes subtle spectral differences among mineral classes and produces accurate, high-confidence predictions in the absence of external distortions. This clean-data benchmark provides a reliable reference for evaluating model robustness and confidence under increasing noise, which is addressed in the following sections. 

\begin{table}[!t]
\renewcommand{\arraystretch}{1.3} 
\caption{Performance evaluation of the HyDeMiC model on clean and noisy hyperspectral data. 
The table presents the MCC, TPR, and PC across different noise levels.}
\label{table_performance}
\centering
\begin{tabular}{|c|c|c|c|c|}
\hline
\textbf{Noise Level (\%)} & \textbf{MCC} & \textbf{TPR} & \textbf{Mean PC(\%)} & \textbf{Median PC(\%)}\\
\hline
\textbf{0 (Clean)} & 1.00 & 1.00 & 99.12 & 99.61\\
\textbf{1} & 1.00 & 1.00 & 99.12 & 99.67\\
\textbf{2} & 1.00 & 1.00 & 99.12 & 99.71\\
\textbf{5} & 0.99 & 0.99 & 99.00  & 99.69\\
\textbf{10} & 0.92 & 0.92 & 95.68 & 97.37\\
\hline
\end{tabular}
\end{table}

\subsection{HyDeMiC prediction on noisy hyperspectral images}
To evaluate HyDeMiC’s robustness, varying levels of noise (1\%, 2\%, 5\%, and 10\%) were introduced into the synthetic 2D hyperspectral dataset (Figure \ref{fig:synthetic_data_generation}c). 
At low noise levels (1\% and 2\% ), the HyDeMiC model retained perfect classification with respect to MCC (MCC = 1.00), demonstrating its ability to effectively handle minor noise without performance degradation (Table \ref{table_performance}). 
TPR remained at 1.00, indicating that the model correctly classified all mineral spectra despite the introduction of minimal noise. 
PC was also consistently high, with a mean PC ranging from 99.12\% to 99.71\% and a median PC from 99.61\% to 99.71\%, see Table \ref{table_performance}.
%

The HyDeMiC maintained near-perfect classification as noise levels increased to 5\%, demonstrating strong robustness to moderate spectral perturbations. As shown in Table~\ref{table_performance}, the model achieves an MCC of 0.999 and a TPR of 0.999, indicating minimal degradation compared to the noise-free case.

\begin{figure*}[!t]
\centering
    \includegraphics[width=\textwidth]{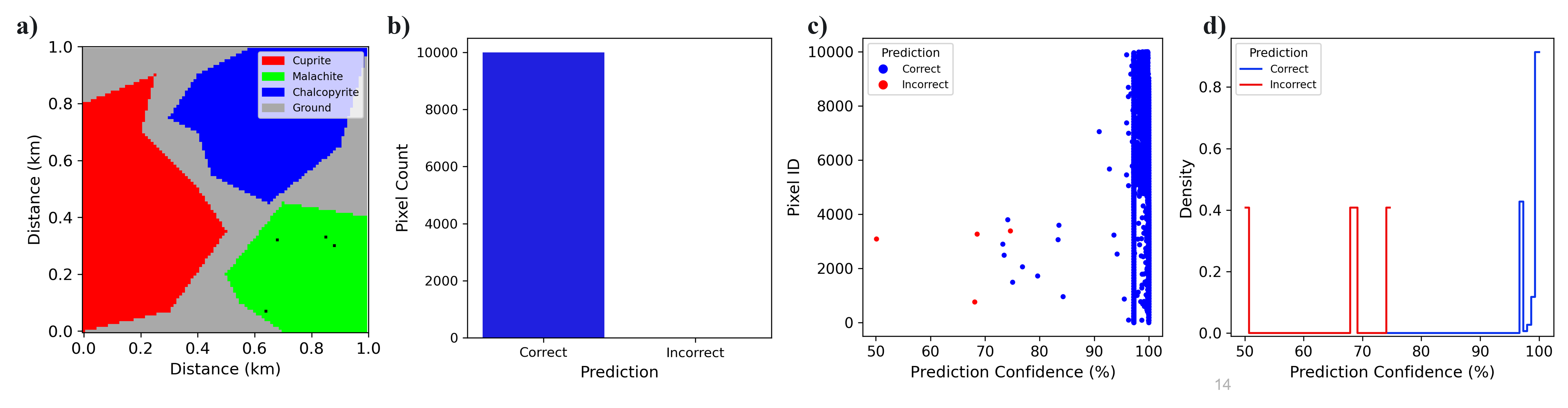}
\caption{
    HyDeMiC prediction results on 5\% noisy synthetic hyperspectral data.
    (a) Spatial distribution of mineral classes predicted by HyDeMiC, 
    (b) Pixel-wise classification summary showing the total number of correct and incorrect predictions,
    (c) Prediction confidence as a function of pixel index, where each point represents an individual pixel colored by correctness, and
    (d) Density-normalized histogram of prediction confidence.
    }
\label{fig:prediction_on_5_noisy_data}
\end{figure*}

The HyDeMiC-predicted mineral distribution (Figure~\ref{fig:prediction_on_5_noisy_data}a) closely matches the true mineral layout. Only four misclassified pixels (highlighted as dark points) are observed among 10,000, resulting in a negligible error rate of 0.04\%. These misclassifications occur mainly near transitions between mineral spectral signatures, where spectral mixing and noise effects are most pronounced.

The pixel-wise classification summary in Figure~\ref{fig:prediction_on_5_noisy_data}b confirms this result, showing an overwhelming majority of correctly classified pixels and very few errors. Figure~\ref{fig:prediction_on_5_noisy_data}c shows that most pixels have confidence values above 95\%, with lower-confidence predictions limited to misclassified boundary pixels.

The density-normalized confidence distributions in Figure~\ref{fig:prediction_on_5_noisy_data}d reinforce these findings. Confidence for correct predictions is sharply peaked near 100\%, while incorrect predictions form a small, low-density tail at lower confidence levels. The mean and median confidences remain high at 99.00\% and 99.69\%, respectively (Table~\ref{table_performance}), indicating that HyDeMiC preserves both accuracy and well-calibrated confidence under moderate noise.


At a 10\% noise level, HyDeMiC maintains strong classification performance despite significant spectral distortion. As shown in Table~\ref{table_performance}, the model achieves an MCC of 0.92 and a TPR of 0.92, indicating that most mineral pixels are correctly identified even under challenging conditions.

\begin{figure*}[!t]
\centering
    \includegraphics[width=\textwidth]{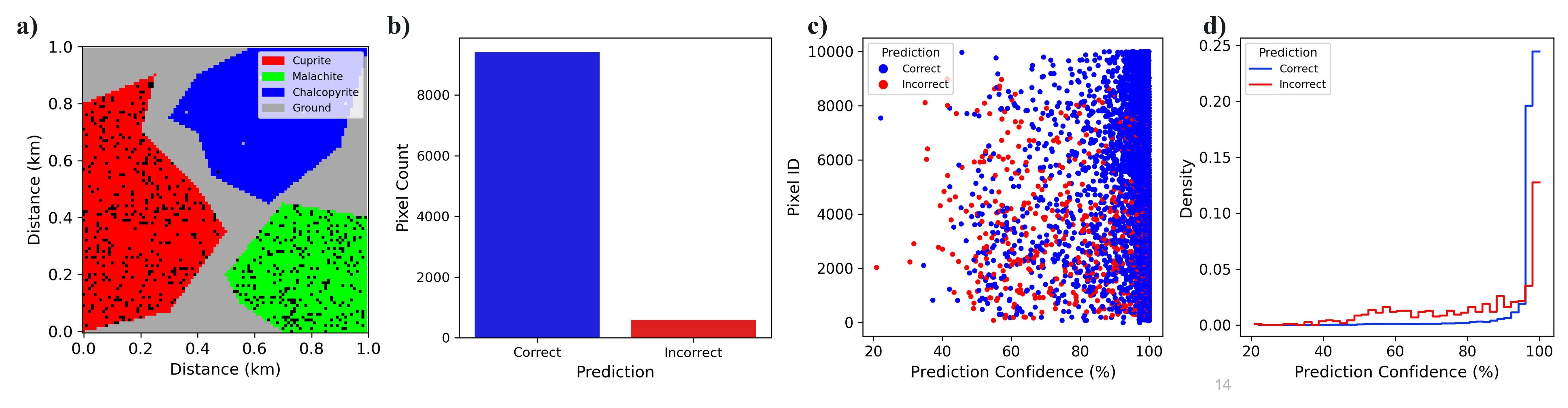}
\caption{
    HyDeMiC prediction results on 10\% noisy synthetic hyperspectral data.
    (a) Spatial distribution of mineral classes predicted by HyDeMiC, 
    (b) Pixel-wise classification summary showing the total number of correct and incorrect predictions,
    (c) Prediction confidence as a function of pixel index, where each point represents an individual pixel colored by correctness, and
    (d) Density-normalized histogram of prediction confidence.
    }
\label{fig:prediction_on_10_noisy_data}
\end{figure*}

A spatial predicted mineral distribution (Figure~\ref{fig:prediction_on_10_noisy_data}a) shows that misclassifications remain limited and spatially structured. Of 10,000 pixels, 591 are incorrectly classified, resulting in a 5.9\% misclassification rate. Most errors occur near transitions between mineral spectral signatures, where spectral mixing and noise amplification are most pronounced, rather than within spectrally homogeneous mineral interiors.

The pixel-wise classification summary in Figure~\ref{fig:prediction_on_10_noisy_data}b confirms that correct predictions remain dominant, despite an increase in incorrect pixels at higher noise levels. Figure~\ref{fig:prediction_on_10_noisy_data}c shows that correctly classified pixels cluster at high confidence values, while misclassified pixels are linked to lower confidence scores.

This trend is evident in the density-normalized confidence distributions in Figure~\ref{fig:prediction_on_10_noisy_data}d. Correct predictions show a strong density peak near 100\% confidence, while incorrect predictions form a broader, lower-density distribution at reduced confidence values. Although overlap between correct and incorrect confidence distributions increases compared to the clean and 5\% noise cases, the mean and median confidence values remain high at 95.68\% and 97.37\%, respectively (Table~\ref{table_performance}). This indicates that HyDeMiC maintains meaningful confidence separation even under severe noise.

These findings demonstrate HyDeMiC’s robustness for classification across diverse mineral groups using hyperspectral data. Even under noisy conditions, the model maintains high accuracy and confidence, highlighting its resilience to spectral distortions. This performance indicates that HyDeMiC can be reliably applied in practical mineral mapping and exploration scenarios where noise and measurement uncertainty are unavoidable.

Beyond mineral-specific performance, the results indicate that HyDeMiC learns transferable spectral representations that are not confined to a single mineral group. The combination of 1D spectral encoding, pixel-wise inference on 2D hyperspectral imagery, and confidence-aware predictions suggests that the framework is inherently suited for broader hyperspectral classification problems involving heterogeneous materials and sensor-induced distortions. These characteristics align with requirements encountered in applications such as lithologic unit identification, soil and sediment property mapping, vegetation and surface material characterization, where reliable pixel-level classification and uncertainty awareness are critical.

%% file: sections/s4_conclusion.tex
\section{Conclusion}
The developed HyDeMiC is a hyperspectral deep learning framework designed for robust mineral classification under varying noise conditions. 
HyDeMiC employs a 1D convolutional neural network trained on spectral signatures of 115 mineral classes to learn discriminative features and to generalize from 1D to 2D hyperspectral imagery via pixel-wise inference.

On clean hyperspectral data, HyDeMiC achieved perfect classification, with MCC and TPR scores of 1.00 and a prediction confidence of nearly 100\%, demonstrating high accuracy and strong calibration.
The model remained robust as noise levels increased.
At low and moderate noise levels, HyDeMiC retained near-perfect performance, with only minimal misclassification and consistently high confidence values. 
Even at high noise levels, where spectral interference alters reflectance patterns, HyDeMiC continued to generalize across mineral groups, maintaining high MCC, TPR, and confidence scores despite a modest decline in accuracy.

The incorporation of confidence-based diagnostics, including density-normalized confidence distributions, provided additional insight into model reliability beyond standard accuracy metrics (both correct and incorrect predictions), highlighting HyDeMiC’s ability to appropriately express uncertainty as noise increases. 
This capability is essential for operational hyperspectral applications, where decision-making depends on both classification outcomes and associated confidence levels.

Overall, the results show that HyDeMiC integrates accurate spectral discrimination, robustness to noise, and interpretable confidence estimation within a unified learning framework. This combination enables reliable hyperspectral mineral mapping under realistic sensing conditions and highlights the framework’s suitability for practical remote sensing and geoscientific applications. Future work will extend the approach to real airborne and satellite hyperspectral datasets, address mixed-pixel effects, and explore uncertainty-aware extensions for large-scale deployment.

Although HyDeMiC is demonstrated for mineral mapping, the same 1D CNN–based encoder–decoder workflow is broadly applicable to hyperspectral classification tasks where materials exhibit diagnostic spectral signatures and predictions are performed pixel-wise on 2D or 3D imagery. Beyond critical minerals, potential applications include lithologic unit discrimination, soil and sediment property mapping, vegetation functional trait or crop-stress classification, surface contamination screening, and infrastructure material identification (e.g., roofing, asphalt, and concrete) using airborne or satellite hyperspectral data.